\documentclass[letterpaper, 10 pt, conference]{ieeeconf}
\usepackage{times}

\IEEEoverridecommandlockouts 

\usepackage[pdftex]{graphicx}
\graphicspath{{figures/}}
\DeclareGraphicsExtensions{.pdf,.jpeg,.png}

\usepackage{graphics} 
\usepackage{epsfig} 
\usepackage{mathptmx} 
\usepackage{times} 
\usepackage{amsmath} 
\usepackage{amssymb}  
\usepackage{booktabs}
\usepackage{float}

\usepackage{helvet}         
\usepackage{courier}        
\usepackage{type1cm}        
\usepackage{makeidx}         
\usepackage{graphicx}        
\usepackage{multicol}        
\usepackage[bottom]{footmisc}
\usepackage{multicol}
\usepackage{multirow}
\usepackage{bm}
\usepackage{color} 
\usepackage[bookmarks=true]{hyperref} 
\usepackage[utf8]{inputenc}
\usepackage[ruled]{algorithm2e}
\usepackage[font=small]{caption}
\usepackage{subcaption}
\usepackage[table]{xcolor}
\usepackage[export]{adjustbox}
\usepackage{threeparttable}

\definecolor{red}{rgb}{1,0,0}
\definecolor{green}{rgb}{0,0.6000, 0.1608}

\definecolor{r1}{rgb}{0,1.0, 0}
\definecolor{r2}{rgb}{0,0,1.0}
\definecolor{r3}{rgb}{0.8627, 1.0000, 0}
\definecolor{r4}{rgb}{1.0000, 0, 0}
\definecolor{r5}{rgb}{0.1176, 0.3922, 0.3922}
\definecolor{r6}{rgb}{0.6078, 0.0039, 0.3922}
\definecolor{r7}{rgb}{0.1961, 0.3922, 0.0392}
\definecolor{r8}{rgb}{0.3137, 0.1961, 0.2745}
\definecolor{r9}{rgb}{0.0392, 0.3922, 0.5882}

\definecolor{CommentPMN}{rgb}{0.0,0.7,0.0}
\definecolor{CommentTS}{rgb}{0,0.0,0.7}
\definecolor{CommentLK}{rgb}{0,0.7,0.7}
\definecolor{TODO}{rgb}{1.0,0.0,0.0}

\newcommand{\ignore}[1]{}

\newcommand{\doctitle}{Inferring Road Boundaries Through and Despite Traffic}
\renewcommand{\doctitle}{Online Inference and Detection of Curbs \\ in Partially Occluded Scenes with Sparse LIDAR}

\title{\LARGE \bf \doctitle}
\author{Tarlan Suleymanov \and Lars Kunze \and Paul Newman
 \thanks{Authors are from the Oxford Robotics Institute (ORI), Dept. of Engineering Science, University of Oxford, Oxford, UK. {\tt\small \{tarlan,lars,pnewman\}@robots.ox.ac.uk} }
}%

\renewcommand{\baselinestretch}{0.97}

\usepackage{cuted}

\usepackage{tikz}

\newcommand\copyrighttext{%
  \footnotesize \copyright 2019 IEEE. Personal use of this material is permitted. Permission from IEEE must be obtained for all other uses, in any current or future media, including reprinting/republishing this material for advertising or promotional purposes, creating new collective works, for resale or redistribution to servers or lists, or reuse of any copyrighted component of this work in other works.}
 
\newcommand\copyrighttextA{%
  \footnotesize Accepted at the 22nd IEEE Intelligent Transportation Systems Conference (ITSC19), October, 2019, Auckland, New Zealand}

\newcommand\copyrightnotice{%
\begin{tikzpicture}[remember picture,overlay]
\node[anchor=south,yshift=10pt] at (current page.south) {\fbox{\parbox{\dimexpr\textwidth-\fboxsep-\fboxrule\relax}{\copyrighttext}}};
\end{tikzpicture}%
}

\newcommand\copyrightnoticeA{%
\begin{tikzpicture}[remember picture,overlay]
\node[anchor=north,yshift=-10pt] at (current page.north) {\fbox{\parbox{\dimexpr\textwidth-\fboxsep-\fboxrule\relax}{\centering\copyrighttextA}}};
\end{tikzpicture}%
}

\begin{document}

\maketitle
\thispagestyle{empty}
\pagestyle{empty}

\setlength{\fboxrule}{0pt}
\copyrightnoticeA
\setlength{\fboxrule}{0.5pt}
\copyrightnotice

\begin{abstract}
Road boundaries, or curbs, provide autonomous vehicles with essential information when interpreting road scenes and generating behaviour plans. 
Although curbs convey important information, they are difficult to detect in complex urban environments (in particular in comparison to other elements of the road such as traffic signs and road markings). These difficulties arise from occlusions by other traffic participants as well as changing lighting and/or weather conditions. Moreover, road boundaries have  various  shapes,  colours  and  structures while motion planning algorithms require accurate and precise metric information in real-time to generate their plans.

In this paper, we present a real-time LIDAR-based approach for accurate curb detection around the vehicle (360 degree). Our approach deals with both occlusions from traffic and changing environmental conditions. To this end, we project 3D LIDAR pointcloud data into 2D bird's-eye view images (akin to Inverse Perspective Mapping). These images are then processed by trained deep networks to infer both visible and occluded road boundaries. Finally, a post-processing step filters detected curb segments and tracks them over time. Experimental results demonstrate the effectiveness of the proposed approach on real-world driving data. Hence, we believe that our LIDAR-based approach provides an efficient and effective way to detect visible and occluded curbs around the vehicles in challenging driving scenarios.
\end{abstract}

\section{Introduction}

Autonomous vehicles are required to detect road boundaries (or curbs) for understanding their surroundings \cite{2018ITSC_kunze} and for generating behaviour plans.
Road boundaries separate drivable road areas from non-drivable areas in the environment. Knowing the boundaries of a road is paramount for many applications, such as autonomous parking, navigation, mapping and path planning. Road boundaries can be used for lateral guidance as part of Advanced Driver Assistant Systems (ADAS), e.g. during parking. However, road boundaries have various shapes, colours and structures, which makes road boundary detection a challenging task. Moreover, road boundaries are often occluded by other traffic participants which makes their detection even more challenging. 

In this work, our goal is to address the problem of road boundary detection using methods of deep learning. Recent advances in deep learning have shown that neural networks can be used for various tasks in robotics, such as segmentation and object detection, which play important roles in solving challenging tasks for operation of self-driving cars. Here we use deep networks to detect visible and infer occluded road boundaries around the vehicle (360$^\circ{}$) from bird's-eye view images that are generated from 3D LIDAR pointclouds. 

\begin{figure}[t!]
    \centering
    \includegraphics[width=\columnwidth]{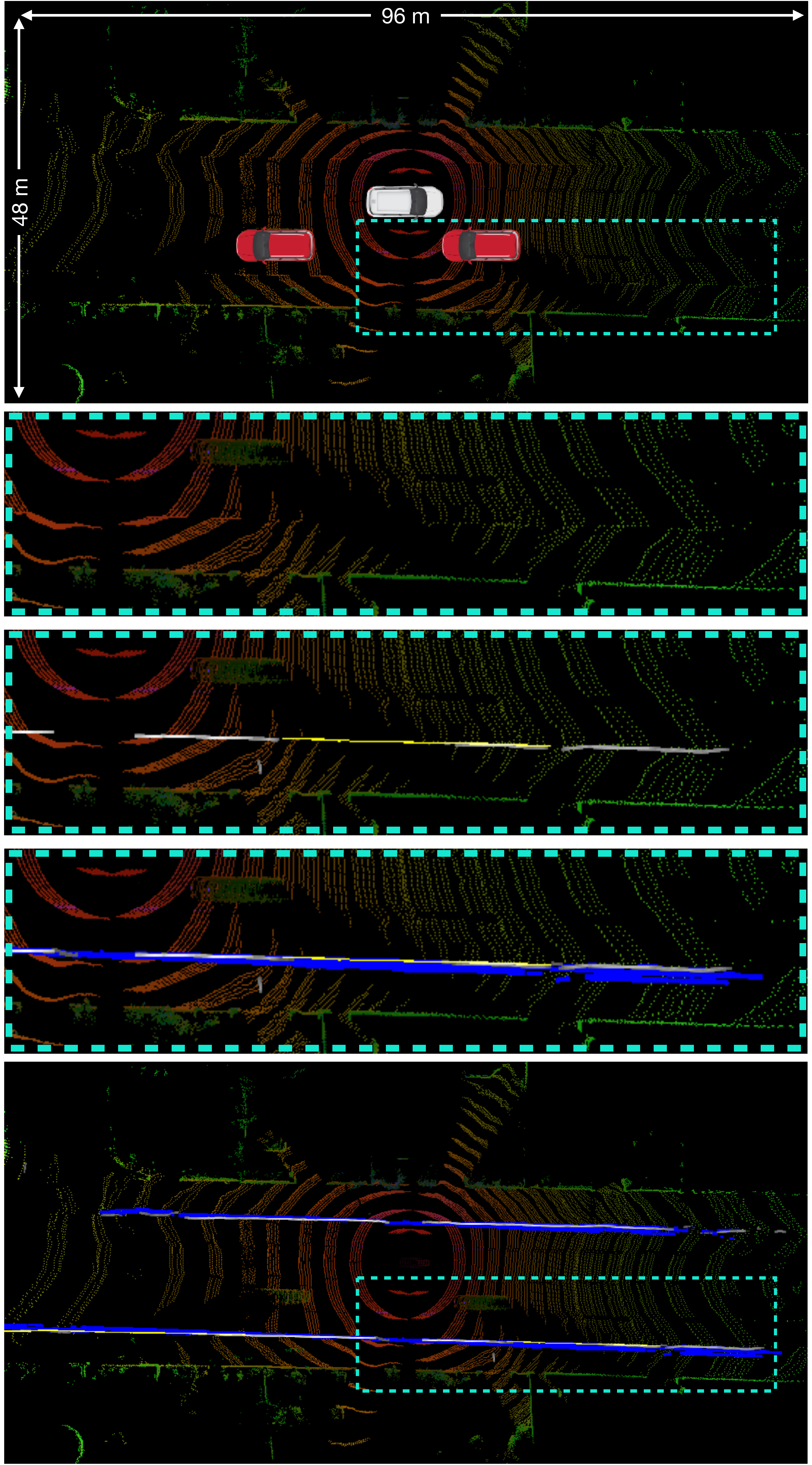}
    \caption{Our 360$^\circ{}$ LIDAR-based curb detection approach. First, LIDAR data from the ego-vehicle (white) is transformed in bird's-eye view images which are then processed by trained deep networks to detect visible (white) and occluded (yellow) curbs. Finally, post-processing steps filters out outliers and tracks curbs over time (blue). The result is a robust curb detection around the vehicle over a total distance of 96 metres.}
    \label{fig:figure_one}
\end{figure}

\begin{figure*}[h!]
    \centering
    \includegraphics[width=\textwidth]{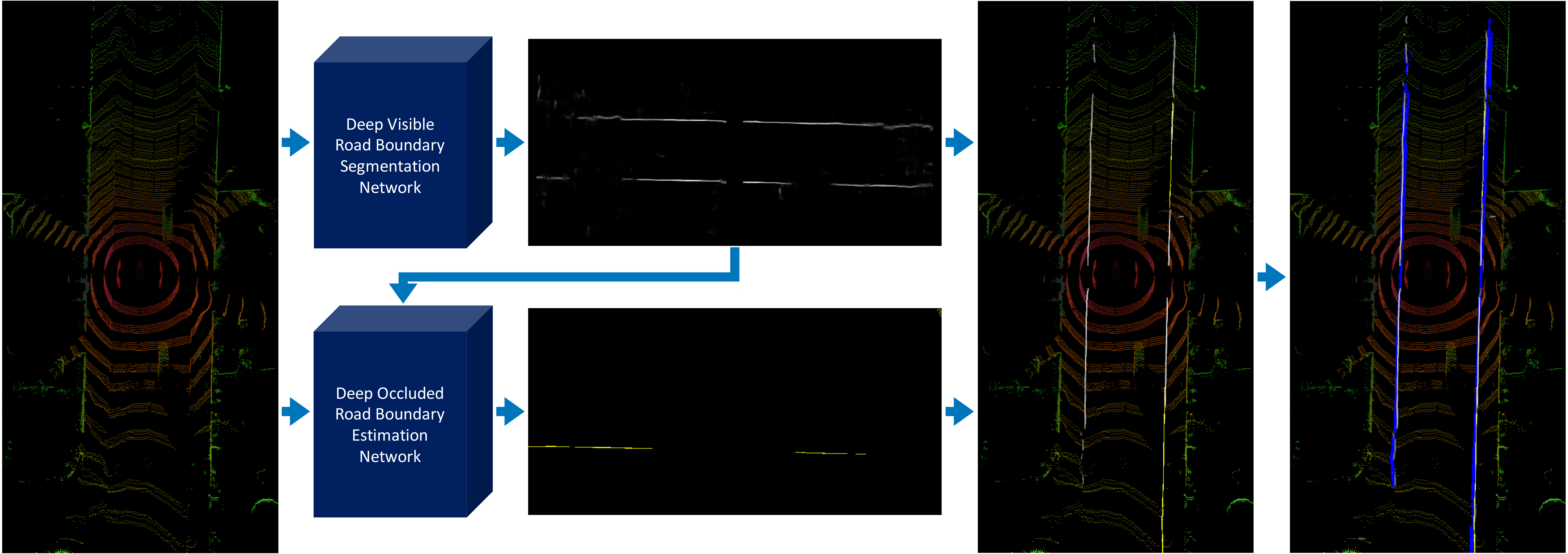}
    \caption{Our 360$^\circ{}$ LIDAR-based  curb  detection approach. A pre-processing step integrates several subsequent laser scans into a coherent coordinate frame and projects them into a bird's-eye view image with height information (left). This image is then processed by two deep segmentation networks to detect both visible and occluded road boundaries (middle). Note that the network responsible for occluded boundaries additionally considers the output of the network for visible boundaries as its input. Finally, the output of both networks is combined and tracked in order to improve the overall performance over a series of processed images (right).}
    \label{fig:pipeline}
\end{figure*}

To this end, the paper makes the following contributions:

\begin{itemize}
    \item a framework to annotate 3D pointclouds and transform them into a set of bird's-eye view images and {\emph{raw}} road boundary masks;
    \item an automatic method for partitioning/labelling {\emph{raw}} training data into two classes (visible and occluded);
    \item a 360$^\circ{}$, real-time deep learning method to detect visible and occluded road boundaries from bird's-eye images without making any assumptions about structure and shape of the road boundaries; and
    \item an experimental evaluation which provides qualitative and quantitative results of our approach on real-world LIDAR data acquired while driving through Oxford. 
\end{itemize}

This remainder of the paper is structured as follows. In Section~\ref{sec:related_work}, we discuss related work on road boundary detection. An overview of our approach is given in Section~\ref{sec:overview}. In Section~\ref{sec:training_data}, we first explain how we obtained a data set for training our deep segmentation/detection networks before we describe our proposed approach in more detail in Section~\ref{sec:approach}. Finally, we provide a thorough qualitative and quantitative evaluation in  Section~\ref{sec:experimental_results}
before we conclude in Section~\ref{sec:conclusions}.  


\section{Related Work}
\label{sec:related_work}
Work on road boundary detection can be divided into two categories: \emph{camera-based} and \emph{LIDAR-based methods}.  

Most camera-based methods address the problem using stereo cameras and 3D geometry to identify road boundaries in the scene \cite{Prinet2016, Kellner2015curbStereoVision, Wang2016boundaryDetectionStereo, Oniga2008multiFramPersistenceMap, Kellner2014ElevationMappingTechniques, Siegemund2011curbCRF, Enzweiler2013TowardsMulti-cue}. In contrast, our previous work \cite{Suleymanov2018Curb} used only a single monocular camera and deep convolutional neural networks for image processing to detect visible and occluded road boundaries. In this paper, we follow a similar machine-learning approach to infer both visible and occluded road boundaries, but we use LIDAR data as input. We do this without making any assumptions on the structure or shape of the curbs.

LIDAR-based methods often rely on more traditional information engineering techniques. In \cite{Hata2016MultilayerLIDAR}, a ring compression analysis on dense 3D LIDAR data followed by false-positive filters was used to detect curb points. Curb models are estimated using Least Trimmed Squares (LTS) and describe the road shape on occluded curbs. However, the approach mostly considers simple examples where curbs are on both sides of the road. Hence, this method would likely fail in more complex scenarios, such as intersections and/or roads with fully occluded curbs. Work by \cite{Yao2012IntegralLaserPoints} uses range and intensity information from 3D LIDAR to detect visible curbs on elevation data, which fails in the presence of occluding obstacles. Similarly, \cite{Zhang2018RoadSegmentationBased} presents a LIDAR-based method to detect visible curbs using sliding-beam segmentation followed by segment-specific curb detection, but fails to detect curbs behind obstacles.

Our proposed approach combines advantages from both research directions camera-based and LIDAR-based methods. In particular, we obtain highly accurate, 3D data about the world from LIDAR sensors which we process in real-time using deep convolutional neural networks (CNNs). 3D LIDAR data allows us to have a larger view angle than a single camera (here 360$^\circ{}$). At the same time, LIDAR data is not restricted by lighting or weather conditions, which allows us to operate the system under various environmental conditions (e.g. rain and fog). 
However, LIDAR-based methods are very data intensive. To circumvent long processing times, we project 3D LIDAR data (i.e. pointclouds) into 2D bird's-eye view images which we then process using CNNs, similar to \cite{Chen_2017, Caltagirone_2017}. Thereby the network robustly detects road boundaries around the vehicle (360$^\circ{}$) in urban traffic scenarios under different weather conditions. 


\section{LIDAR-based Curb Detection: An Overview}
\label{sec:overview}

In this work, we detect and infer visible and occluded road boundaries based on sparse LIDAR data using deep learning methods.  

LIDAR data is used for both generating training examples and inferring road boundaries. To this end, we propose a novel framework to semi-automatically generate and boost the training data (Section~\ref{sec:training_data}). 
Moreover, we use 3D LIDAR data to detect and infer road boundaries around the vehicle (Sections~\ref{sec:approach}). 

An overview of our $360^\circ{}$ curb detection approach is given in Figure~\ref{fig:pipeline}. 
In a pre-processing step, a sequence of LIDAR scans are integrated and projected into a bird's-eye view image. Using this image, a first network detects visible road boundaries with a fully convolutional neural network and passes the output to a second network. The purpose of the second network is to infer road boundaries given both the original bird's-eye view image and a  mask of detected visible road boundaries. The generated output represents segments of road boundaries in a hybrid, discrete-continuous form. To improve the overall performance, a post-processing step filters out noise by consolidating detections in subsequent images and by tracking detected curbs over time. 

\begin{figure}[t!]
    \centering
    \includegraphics[width=1.0\columnwidth]{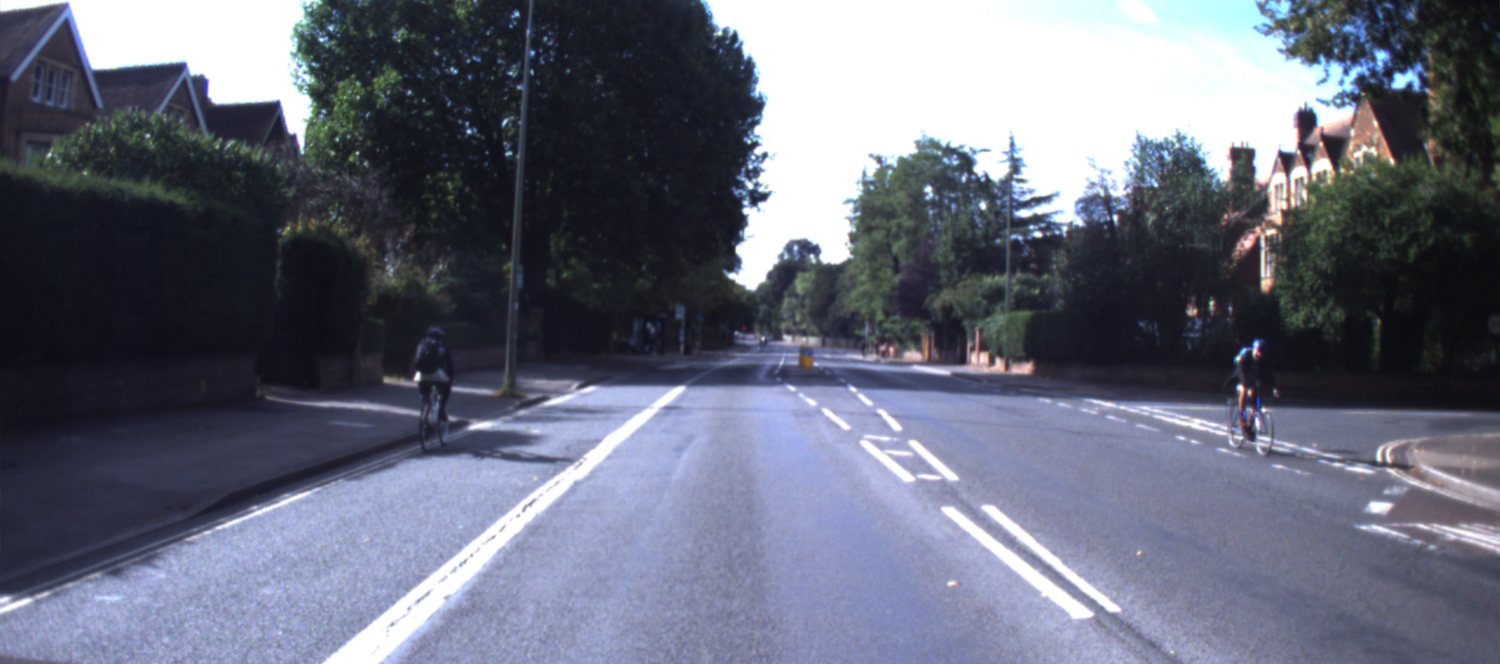}
    
    \vspace{0.1em}
    
    \includegraphics[width=1.0\columnwidth]{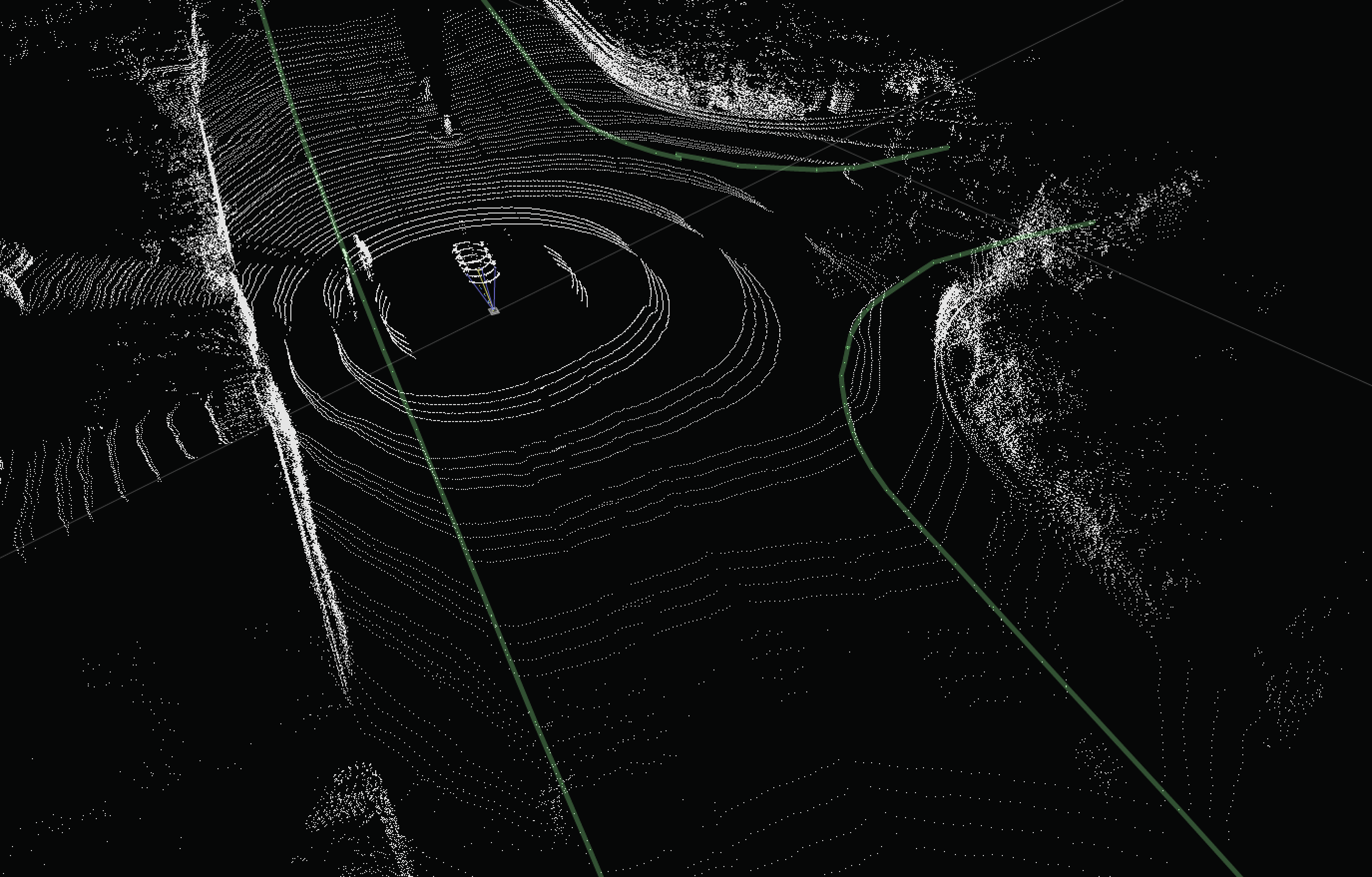}
    \caption{Integrated 3D pointcloud data acquired by driving through Oxford and manually annotated with road boundaries.}
    \label{fig:map_builder}
\end{figure}

\begin{figure}[h]
	\centering
	\begin{subfigure}{0.535\columnwidth}
		\includegraphics[width=\textwidth]{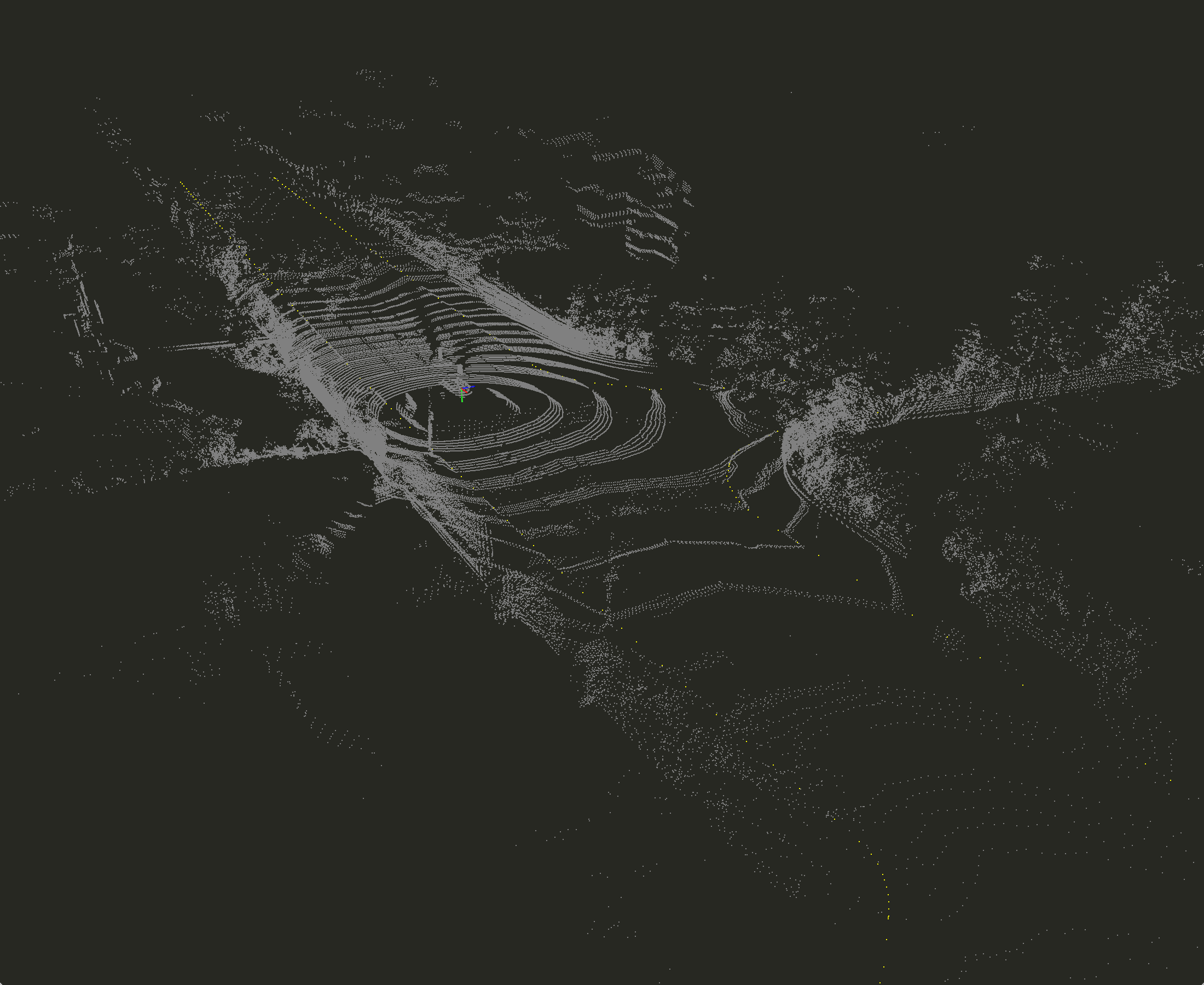}
		
		\vspace{0.1em}
		
		\includegraphics[width=\textwidth]{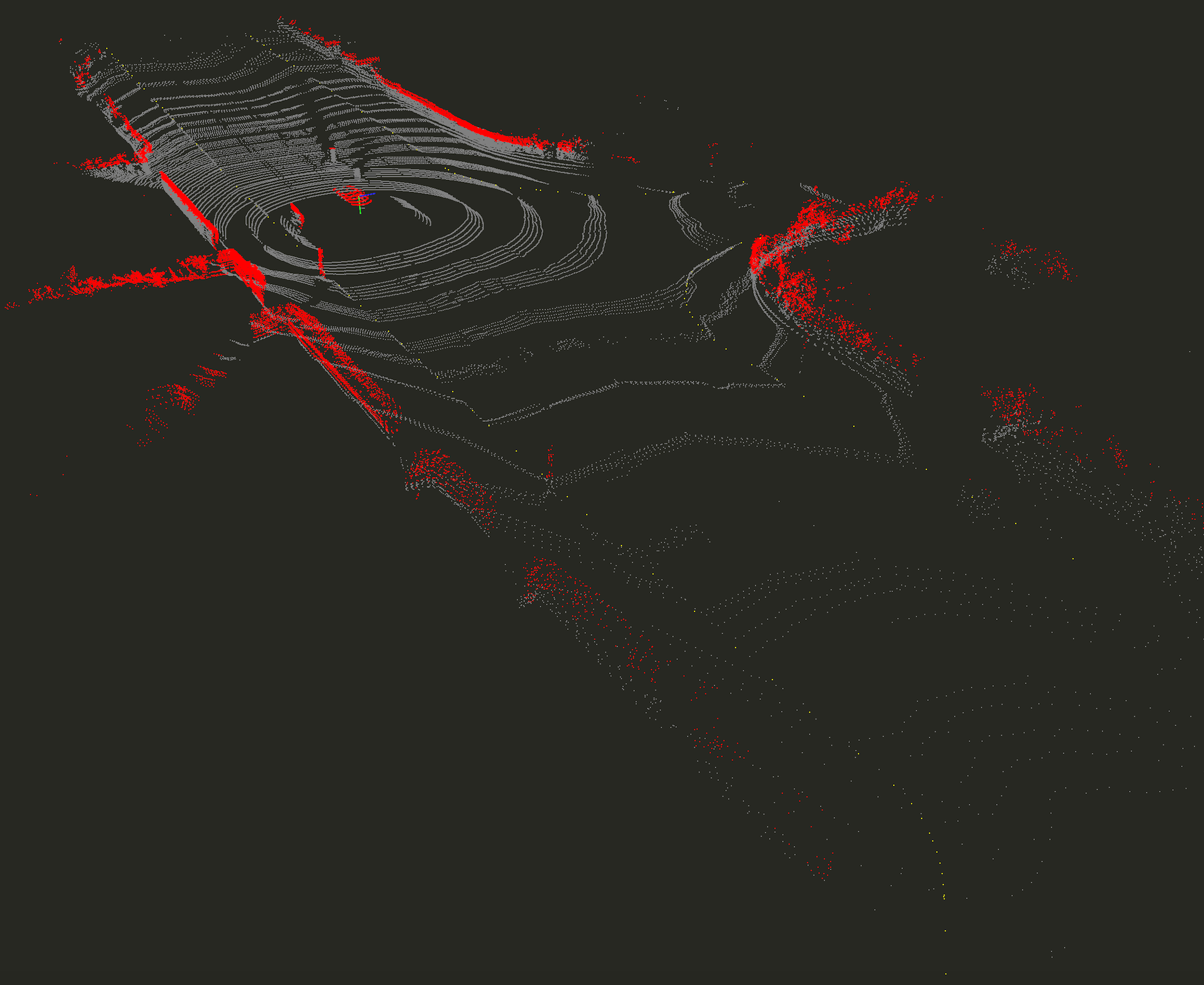}
	\end{subfigure}
	\vspace{0.05em}
	\begin{subfigure}{0.440\columnwidth}
		\includegraphics[width=\textwidth]{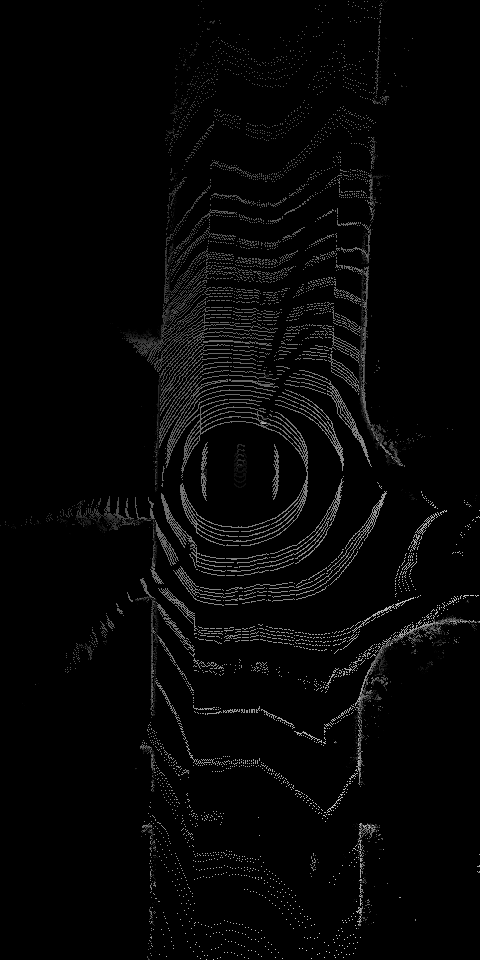}
	\end{subfigure}
	\caption{Top left: integrated LIDAR pointcloud. Bottom left: Filtered and trimmed pointcloud. Right: Projected bird's-eye view image. }
	\label{fig:td_gemeration}
\end{figure}


\section{Obtaining Training Data and Ground Truth}
\label{sec:training_data}

Training deep networks to obtain high-performance and well generalised models requires large datasets with ground truth labels. Moreover, the variability of road boundaries in shape and structure requires training data to include samples from different environments. Although obtaining \emph{raw} data is relatively simple, fine-grained annotation of data requires human interaction and can be extremely time consuming. For example, outlined distinct regions in an image must be associated with corresponding class labels. To obtain ground truth lables we annotated road boundaries in 3D integrated pointcloud data, which was collected by a test car fitted with a 3D LIDAR (Velodyne VLP-32C) on the roof (Figure \ref{fig:map_builder}). 
To register and integrate a set of subsequent LIDAR scans in a coherent coordinate frame, we estimate the vehicle's motion using Visual Odometry (VO)~\cite{DBLP:conf/cvpr/NisterNB04}. In this work, we used images acquired by a Point Grey Bumblebee XB3 camera, mounted on the front of the platform facing towards the direction of motion. In particular, our implementation of VO uses FAST corners~\cite{fast2005} combined with BRIEF descriptors~\cite{brief2012}, RANSAC~\cite{ransac1981} for outlier rejection, and nonlinear least-squares refinement. 

Based on integrated 3D pointclouds we then generate 2D projections which are transformed into bird's-eye view (IPM) images. To do this, we take a laser scan and remove all points above the LIDAR device. As the LIDAR is mounted on top of the car, points above the LIDAR cannot be part of road boundaries. Similarly, we remove points that are below the LIDAR by more than 3.55 metres because some points may appear below the road surface as reflections. Also, we only keep  points if they are located within 24 metres away from the car in x direction and within 48 metres in z direction. Thus we obtain trimmed pointclouds that we transform into bird's-eye view images (Figure \ref{fig:td_gemeration}). In similar way, we obtain image masks by projecting annotations of road boundaries. Note that input bird's-eye view images consist of three channels, range, intensity and height. In this work, we used LIDAR data from the OxfordRobotcar dataset~\cite{RobotCarDatasetIJRR} to  obtain a new data set of bird's-eye view images and semi-annotated (labelled) road boundary masks. 

\begin{figure*}[h!]
    \centering
    \includegraphics[width=0.245\textwidth]{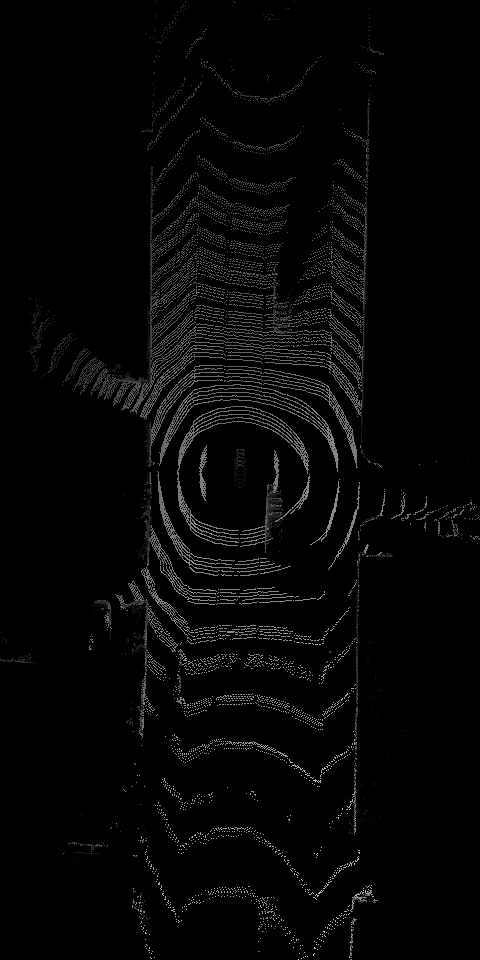}
    \includegraphics[width=0.245\textwidth]{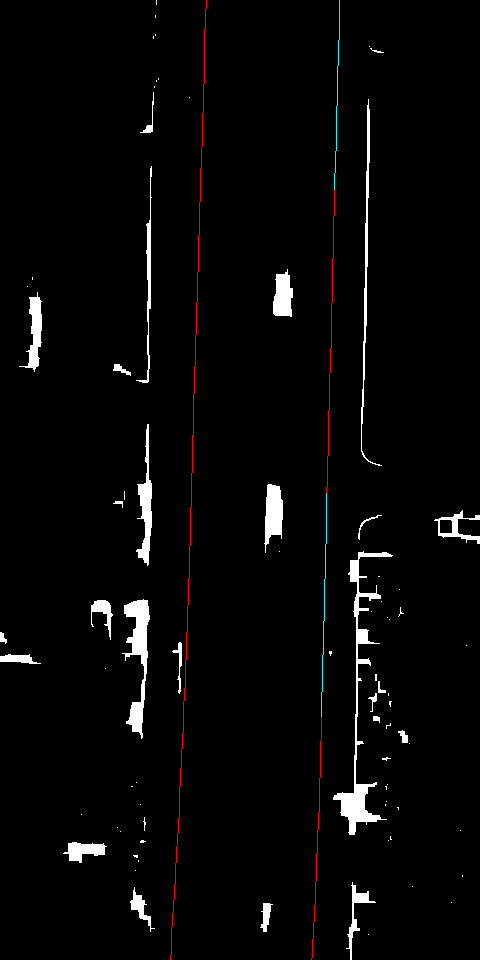}
    \includegraphics[width=0.245\textwidth]{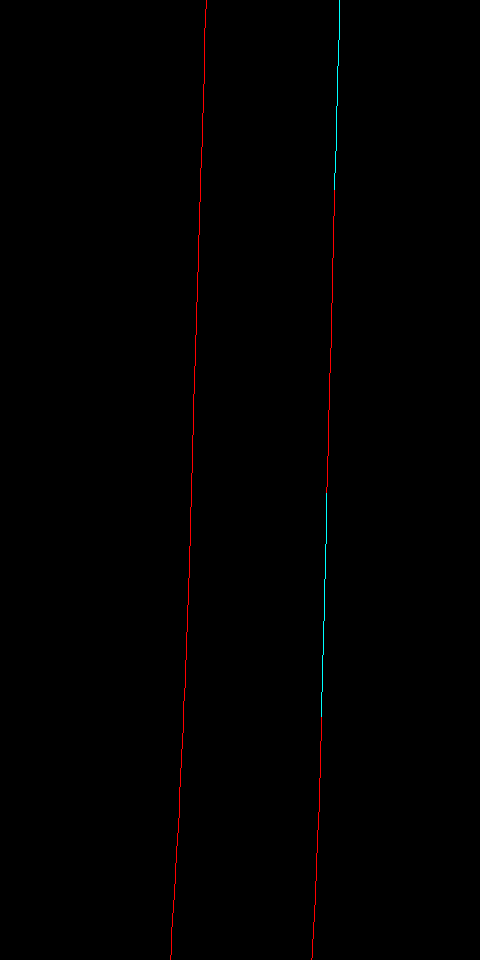}
    \includegraphics[width=0.245\textwidth]{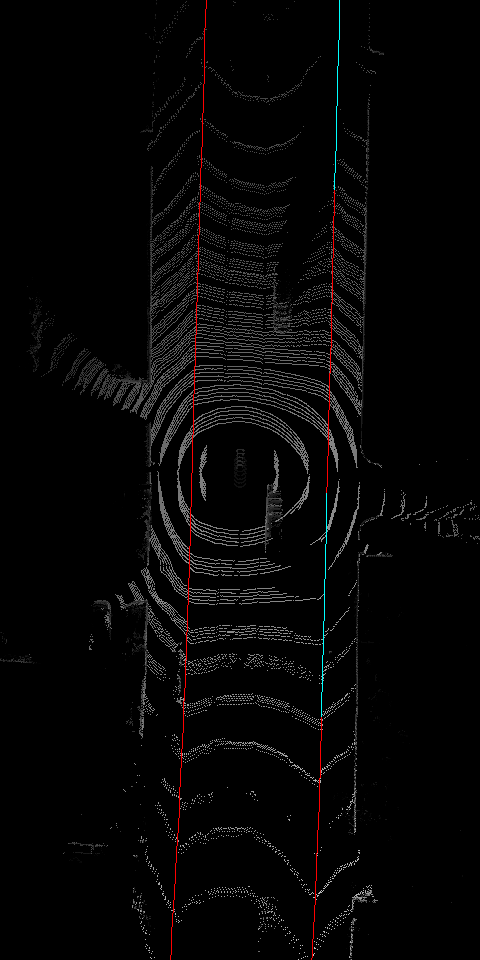}
    \caption{Partitioning of training data. From left to right: bird's-eye view image, detected obstacles as well as visible and occluded curbs, curbs labels only, bird's-eye view image with labels.}
    \label{fig:td_partitioning}
\end{figure*}

\subsection{Partitioning Training Data: Visible and Occluded Curbs}

To detect all road boundaries in a given scene we split the problem into two tasks: first we detect road boundaries that are visible from the laser and then infer road boundaries that are occluded by other road users. To achieve this, we partition the training data into visible and occluded classes. 

However, note  that the obtained  masks contain both visible and occluded road boundaries as a single class as they were generated by projecting 3D annotations into bird's-eye view images. To determine which points are visible and which are occluded we use the hidden point removal operator as described in \cite{Katz:2007:DVP:1276377.1276407}. The operator determines all visible points in a pointcloud when observed from a given viewpoint. This is achieved by extracting all points residing on the convex hull of a transformed pointcloud. These points resemble the visible points, all other (labeled) points are considered as \emph{hidden} (or occluded).   
We take the previously trimmed pointclouds and create binary bird's-eye view images by taking the height of points from the ground into account. The points that are within a predefined height difference from the LIDAR roughly correspond to the points (obstacles) that are blocking the view. By putting together raw labels and binary masks of obstacles, obtained by running the hidden point removal algorithm, we obtain separate masks for visible and occluded road boundaries (Figure \ref{fig:td_partitioning}).

\begin{figure*}[h!]
    \centering
    \includegraphics[width=0.195\textwidth]{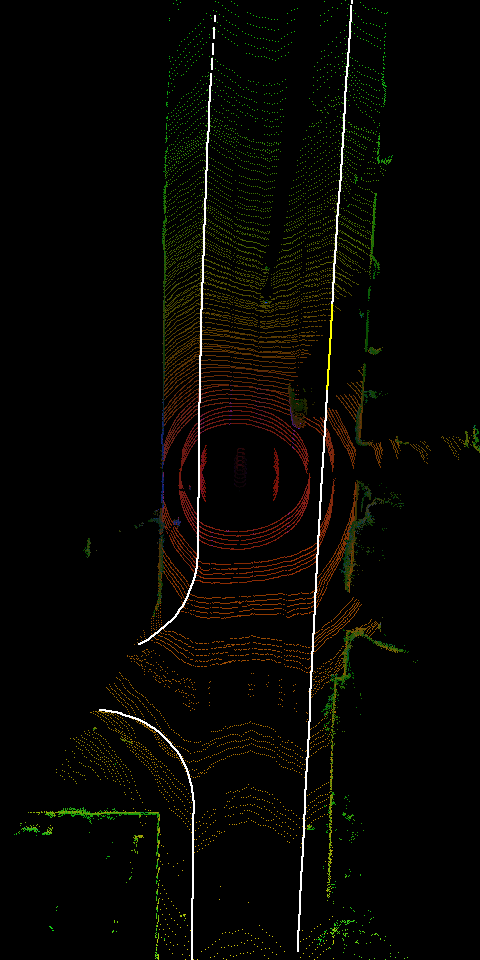}
    \includegraphics[width=0.195\textwidth]{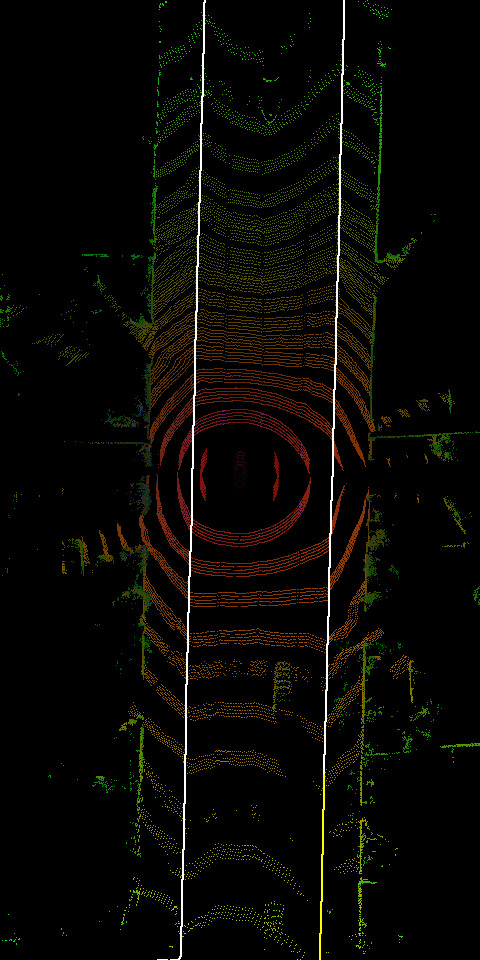}
    \includegraphics[width=0.195\textwidth]{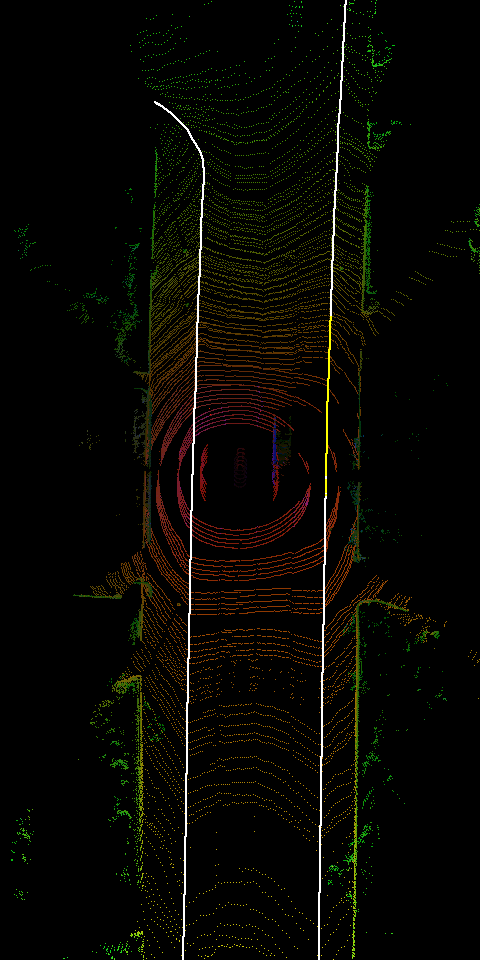}
    \includegraphics[width=0.195\textwidth]{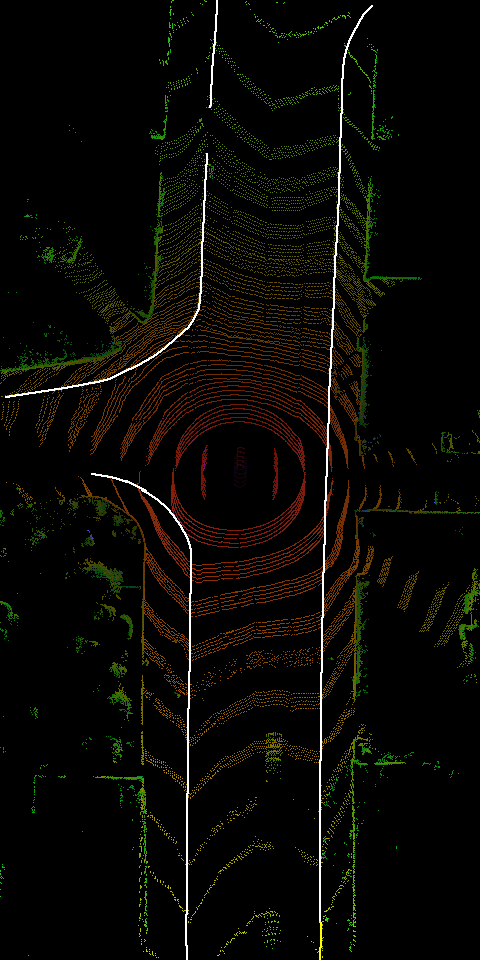}
    \includegraphics[width=0.195\textwidth]{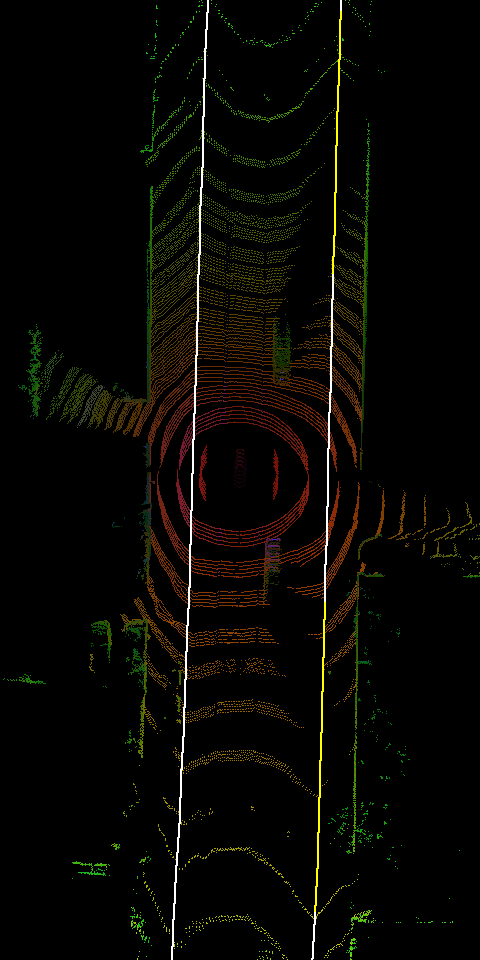}

    \caption{Examples of labelled training data. Visible curbs are marked in white, while occluded curbs are marked in yellow.}
    \label{fig:training_data_examples}
\end{figure*}


\section{Our approach}
\label{sec:approach}

Separating the training data into two classes (visible and occluded curbs) allows us to address the problem in two steps. Note, however, that these steps are linked (see Figure~\ref{fig:pipeline}). In a first step we detect only visible curbs. These detected curbs then provide additional information for the second step, the detection of occluded curbs. In the following we briefly describe the architectures used for both steps.

\subsection{Detecting visible curbs}
\label{sec:visible_detection}

To detect visible curbs we use the U-net architecture~\cite{DBLP:journals/corr/RonnebergerFB15}. 
U-net is a fully convolutional network which concatenates higher resolution “input-side” features from convolution layers with up-sampled outputs form deconvolution layers. This enables the network to detect and segment objects such as curbs more precisely. Although this approach has been successful for visible curbs, it did not generate the desired outcome for occluded curbs. The reasons for this are twofold: first, the network's limited receptive field, which is not big enough to capture context around large obstacles to estimate the position of curbs behind them, and second, the lack of structure (model-free) which prevents the network to infer very thin curves of occluded road boundaries within an image. Hence, we have employed a different approach for the detection of occluded curbs which we explain in the next section.   

\begin{figure}[h!]
    \centering
    \includegraphics[width=\columnwidth]{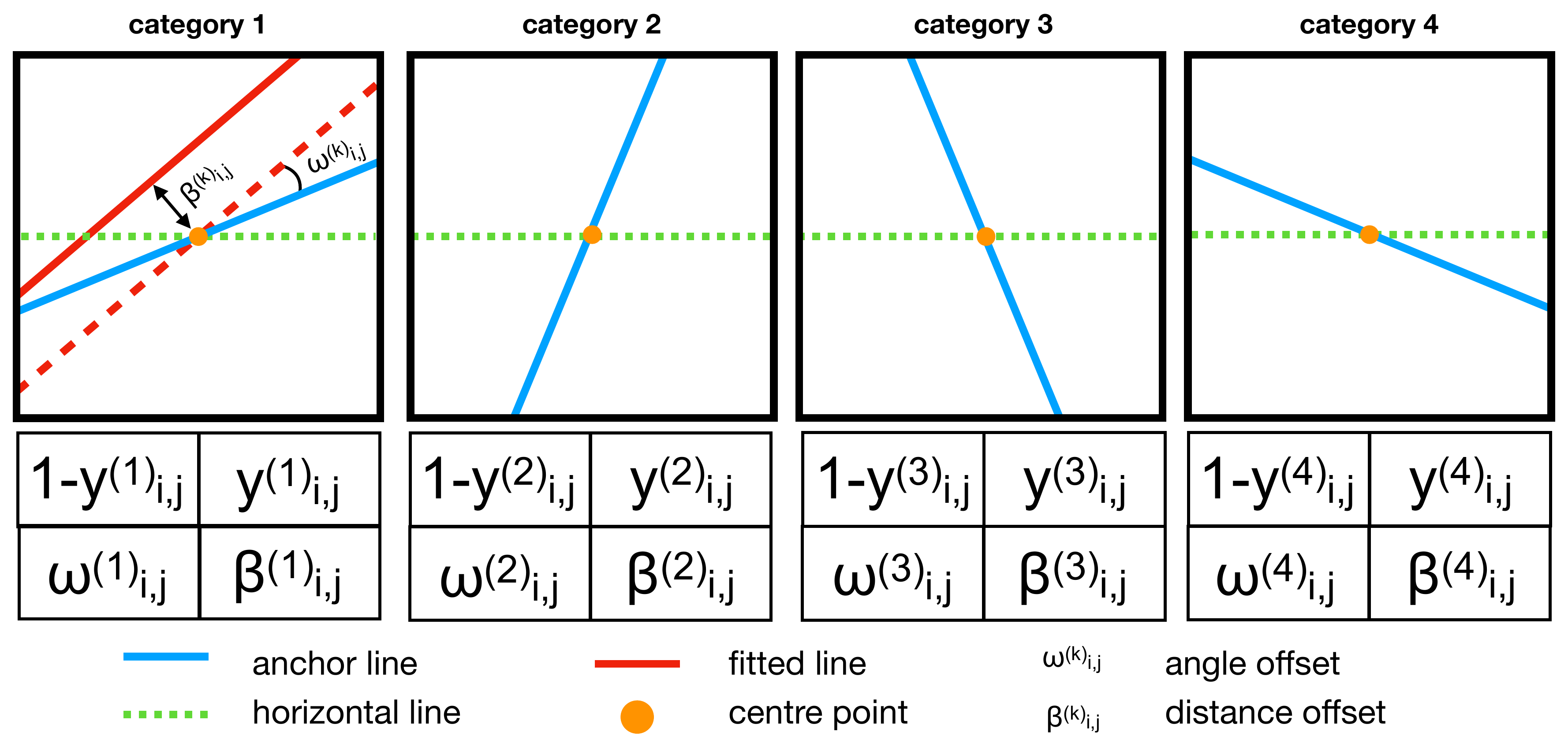}
    \caption{Parameterisation of curb lines in discrete-continuous form. The four categories correspond to the four anchor lines. }
    \label{fig:anchor_lines}
\end{figure}

\subsection{Inferring occluded curbs}
\label{sec:occluded_detection}

In this section, we explain our approach on inferring curbs in partially occluded scenes.

Our model consists of several convolutional layers that produce an output of detected curbs as discrete lines at multiple scales. Instead of pixels, a trained network estimates parameters of lines that correspond to cells in a grid (at different scales). It discretises the output space of lines into a set of default (anchor) lines over different orientation angles. At inference time, the network then generates probabilities for the presence of occluded curbs for each anchor line and its orientation. Making predictions of occluded curbs at multiple scales is important due to the different sizes and shapes of occluding obstacles.

In this work, we have selected three scales of parameterised outputs due to run-time constraints and a given accuracy target. Pixel-wise curb labels are converted to parameterised labels by dividing curb masks (at each scale) into a grid. Lines in each grid cell are parameterised in a discrete-continuous form: first, fitted lines are assigned to one of four types of anchor lines, and secondly, offsets between fitted and anchor lines are calculated. Anchor lines pass through the centre of a grid cell at different angles (22.5$^\circ{}$, 67.5$^\circ{}$, 112.5$^\circ{}$ and 157.5$^\circ{}$). During fitting, lines are assigned to the closest anchor line. Once a fitted line is discretised, two continuous parameters are calculated: (1) an angle offset between a fitted and the respective anchor line ($\omega_{i,j,gt}^{k}$), and (2) a distance from the centre of the cell to the fitted line ($\beta_{i,j,gt}^{k}$). As a result, we obtain 16 numbers for each grid cell, 4 numbers for each line category.

Estimating the presence of a curb line is a classification problem, but estimating adjustments to that line is a regression problem. 
To teach our network to perform the classification and regression at the same time, a discrete-continuous loss is applied during the training process.
The total loss of the model $L_t$ is defined as:

\begin{equation} 
    \label{eq:discret-cont}
    L_t = L_d + \alpha L_c = \sum_{i=1}^{S} L_{d_i} + \alpha \sum_{i=1}^{S} L_{c_i}
\end{equation}

where $L_d$ is a discrete loss of the curb line category classification, $L_c$ is a continuous loss of the curb line parameters' regression, $\alpha$ is a weight term, and $S$ denotes the number of scales (here 3). 

Let $\hat{p}_{i,j}^{k}$ be a softmax output of the network for the $k$-th anchor line category in the $j$-th cell of the $i$-th scale, then the discrete loss for the $i$-th scale is:

\begin{equation} 
    \label{eq:discrete}
    L_{d_i} = -\sum_{j=1}^{C_{i}} \sum_{k=1}^{A} (y_{i,j}^{k} \log (\hat{p}_{i,j}^{k}) + (1 - y_{i,j}^{k}) \log (1 - \hat{p}_{i,j}^{k}) )
\end{equation}
where $A$ is the number of anchor line categories (there are 4 categories), $C_{i}$ is the number of cells in the $i$-th scale and $y_{i,j}^{k}$ is the ground truth for the $k$-th anchor line category in $j$-th cell of the $i$-th scale. 

The continuous loss is a smooth $L1$ loss between the predicted line ($\omega_{i,j,pr}^{k}$, $\beta_{i,j,pr}^{k}$) and the ground truth line ($\omega_{i,j,gt}^{k}$, $\beta_{i,j,gt}^{k}$) parameters. The continuous loss for the $i$-th scale is defined as:
\begin{equation} 
    \begin{aligned}
    \label{eq:continuous}
    L_{c_i} & = \sum_{j=1}^{C_{i}} \sum_{k=1}^{A} (y_{i,j}^{k} (smooth_{L1} (\omega_{i,j,pr}^{k} - \omega_{i,j,gt}^{k}) \\
    & + smooth_{L1} (\beta_{i,j,pr}^{k} - \beta_{i,j,gt}^{k}))
    \end{aligned}
\end{equation}
where $smooth_{L1}$ is define as in \cite{Girshick2015FastRCNN}.

\begin{figure*}[h!]
    \centering
    \includegraphics[width=0.245\textwidth]{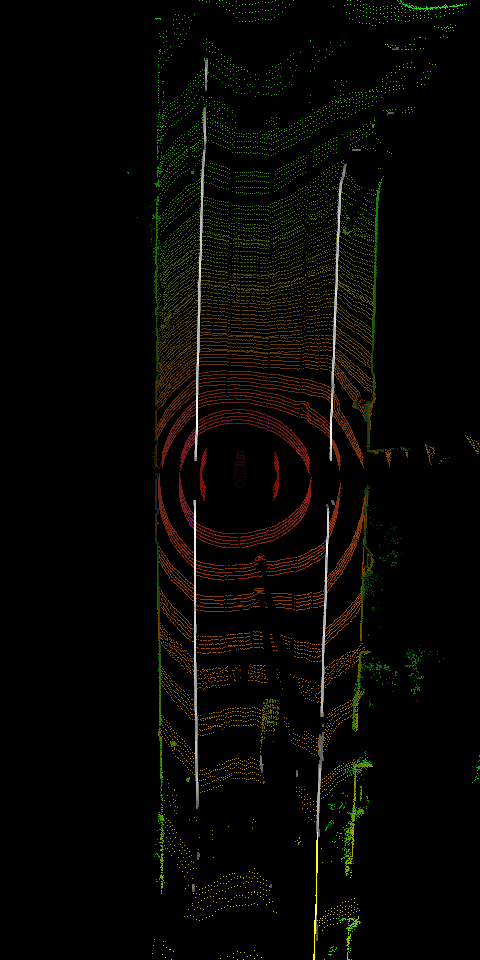}
    \includegraphics[width=0.245\textwidth]{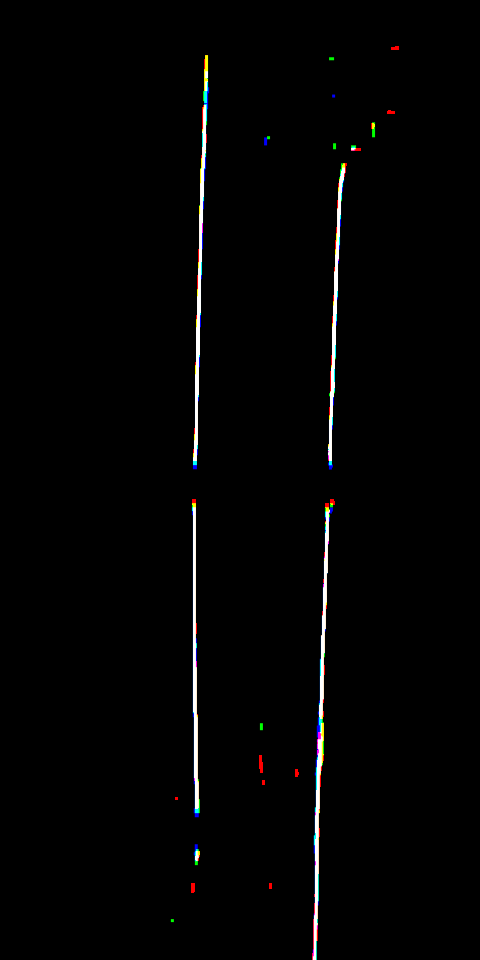}
    \includegraphics[width=0.245\textwidth]{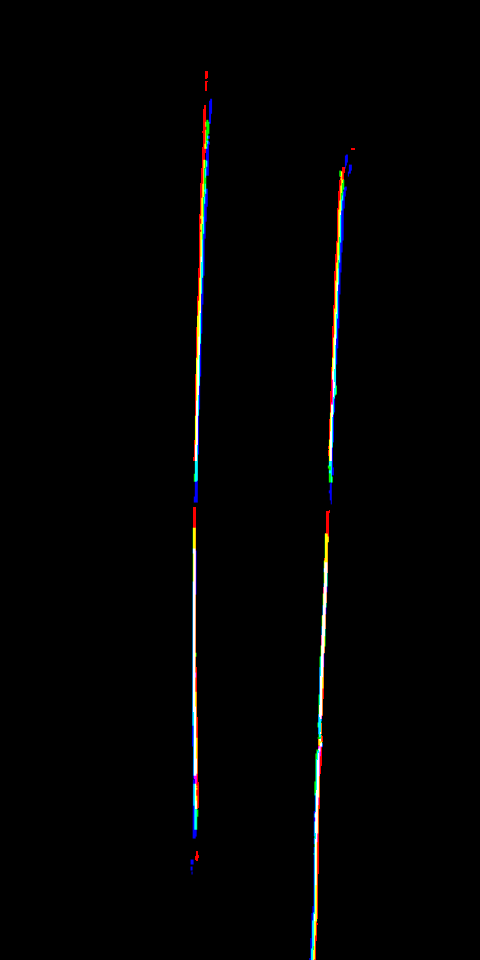}
    \includegraphics[width=0.245\textwidth]{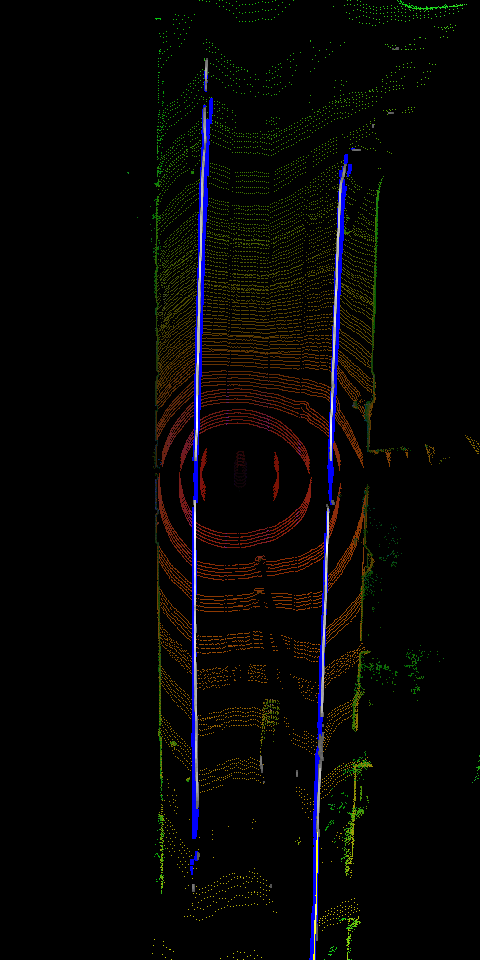}
    \caption{Post-processing. From left to right: Output of curb detection networks (visible and occluded). Filtering step in which noise (highlighted in colour) is removed. Tracking step in which features tracked over time (in colour) are added. Finally, curb detection result after post-processing.}
    \label{fig:post_processing_steps}
\end{figure*}

\begin{figure}[b!]
    \centering
    \includegraphics[width=0.5\columnwidth]{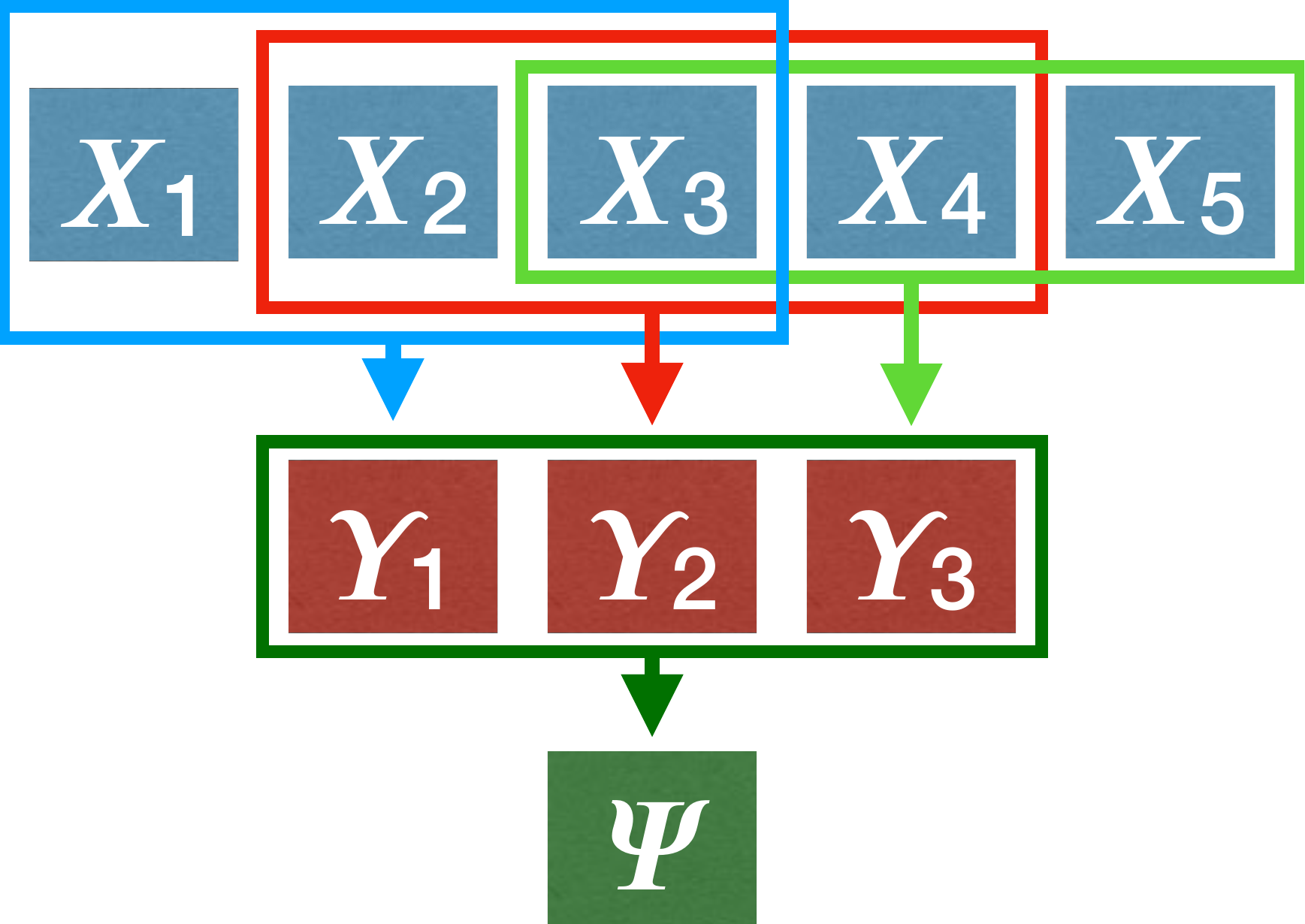}
    \caption{Post-processing steps: A first step consolidates detection results of several subsequent scans ($X_i$) and generates a filtered output ($\Upsilon_i$). A second step, tracks detected segments over a series of filtered images ($\Upsilon_i$) and generates a final output ($\Psi$).}
    \label{fig:post_processing_fig}
\end{figure}

To increase the receptive field of the model we added intra-layer convolutions \cite{pan2017spatialasdeep} before the multi-scale parameter estimation layers.
Traditional layer-by-layer convolutions are applied between feature maps, but intra-layer convolutions are slice-by-slice convolutions within feature maps. 
Hence, intra-layer convolutions capture aspects across the whole image and can thereby capture spatial relationships over longer distances. For example, there is a strong correlation between the length of the occluded curbs and the size of objects which are obstructing the view (ranging from 10-15 pixels through occlusions by traffic cones to 200-300 pixels through occlusions by several parked cars).


\subsection{Post-processing}
\label{sec:post_processing}

Although the generated bird's-eye view images integrate five subsequent laser scans, our models do not take any temporal information into account explicitly. That is, each inference step is independent from the previous one. However, using temporal information can be useful in two ways: filtering out false positives and tracking true positives. To achieve this, we transform inputs at different time steps, $t_l$ and $t_l'$, into a common reference frame using the following transformation:

\begin{equation} 
    \label{eq:transformation}
    T(t_l', t_l) = \left[ 
                        \begin{array}{cc}
                            R(t_l',t_l) & t(t_l', t_l) \\
                            0 & 1
                        \end{array}
                    \right]
\end{equation}

where $R(t_l',t_l)$ is a rotation matrix and $t(t_l', t_l)$ is a translation vector. As we explained in Section~\ref{sec:training_data}, VO is used to estimate the vehicle's ego-motion that allows us to obtain transformations between the vehicle's poses at different time steps. However, VO provides transformations corresponding to a camera frame rate. To obtain transformations between successive laser scans we use interpolation and calculate the transformation between time frames $t_l$ and $t_l'$ as follows \cite{Aldera2019FastRadar}:

\begin{align} 
    \label{eq:chain_transformation}
    T(t_l', t_l) & = T(t_l', t_{vo}') \cdot T(t_{vo}', t_{vo}) \cdot T(t_{vo}, t_l) \\
    s.t. & \ t_{vo}' \leq t_l',  \ \ t_{vo} \geq t_l \nonumber
\end{align}

where $t_{vo}'$ and $t_{vo}$ are the closest time steps of the VO with respect to laser frames and where $T(t_{vo}', t_{vo})$ is defined as follows:

\begin{equation}
    T(t_{vo}', t_{vo}) = \prod_{i=t_{vo}'}^{t_{vo}-1}T(i, i+1)
\end{equation}

To obtain $T(t_l', t_{vo}')$ and $T(t_{vo}, t_l)$ we interpolate in $[t_{vo}', t_{vo}'+ 1]$ and $[t_{vo}-1, t_{vo}]$ respectively. Using these transformations we apply two post-processing steps as follows:

\textbf{Filtering}. In the first step, we transform the last three output masks of detected road boundaries into a common reference attached to the current frame. Then we construct a histogram of output mask size (480x960) by counting the number of overlapping pixels with a value grater than threshold of 0.7 (which was determined experimentally). We keep the detected road boundaries that appear in all three frames above the threshold and disregard the rest. As a result noise is filtered out from the output mask as shown on histogram in Figure \ref{fig:post_processing_steps}.

\textbf{Tracking}. In the second step, we perform a similar procedure as outlined above. However, this time we consider road boundary masks from the last three frames that were generated by the first step (as shown in Figure \ref{fig:post_processing_fig}). By taking the union of these masks we track the detected road boundaries over the time. Integrating temporal information helps to close gaps between boundary segments (Figure \ref{fig:post_processing_steps}). 

\begin{figure*}[h!]
    \centering
    \includegraphics[width=0.195\textwidth]{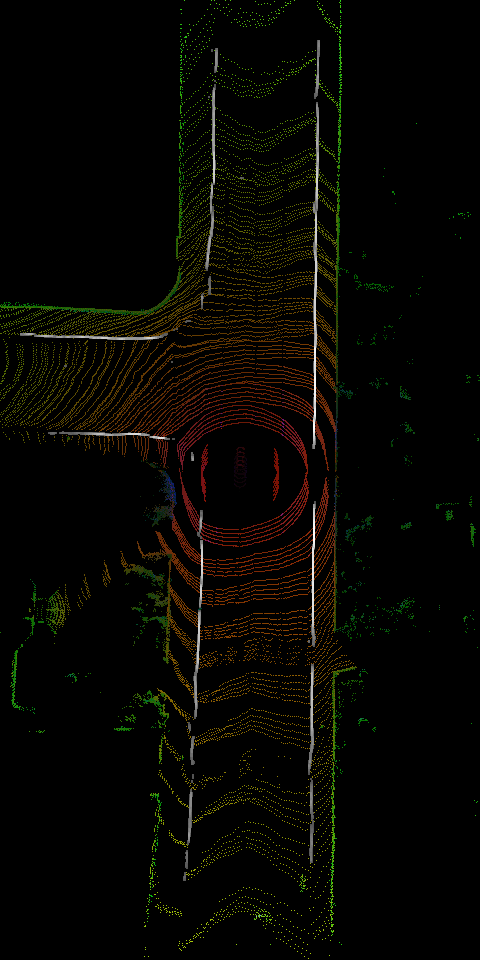}
    \includegraphics[width=0.195\textwidth]{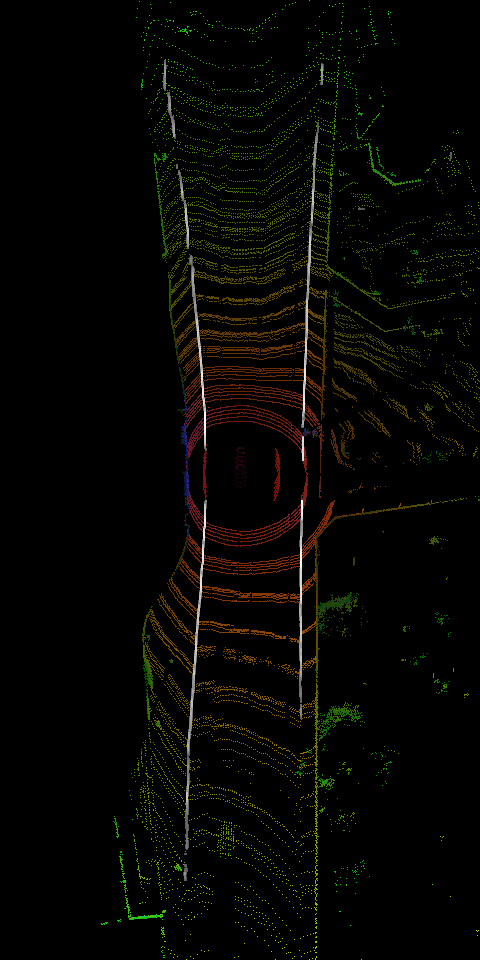}
    \includegraphics[width=0.195\textwidth]{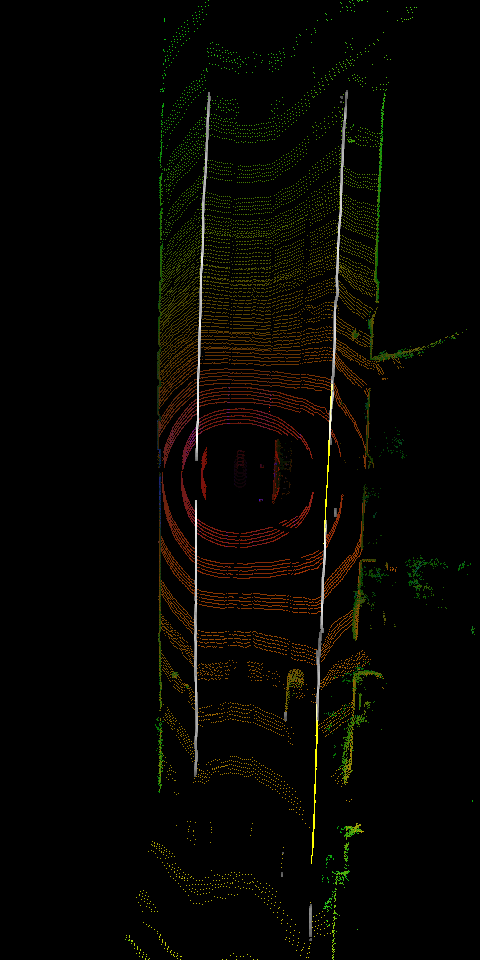}
    \includegraphics[width=0.195\textwidth]{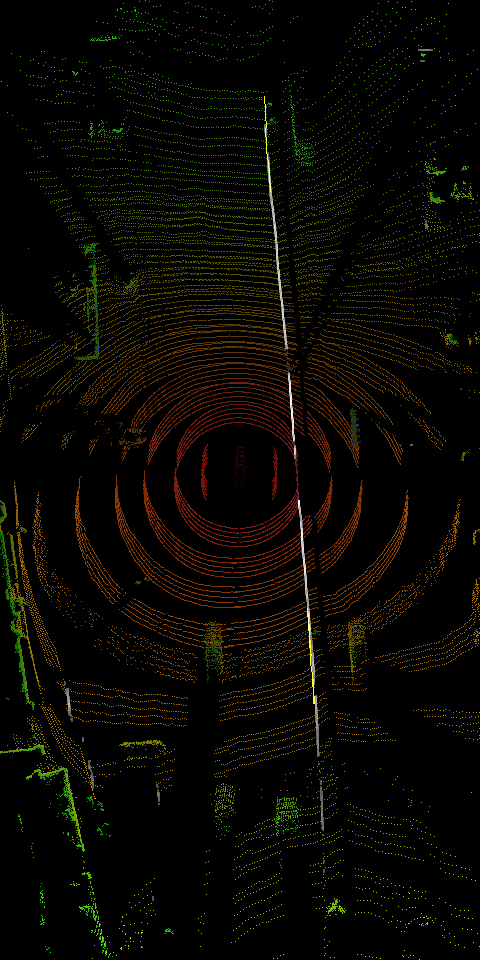}
    \includegraphics[width=0.195\textwidth]{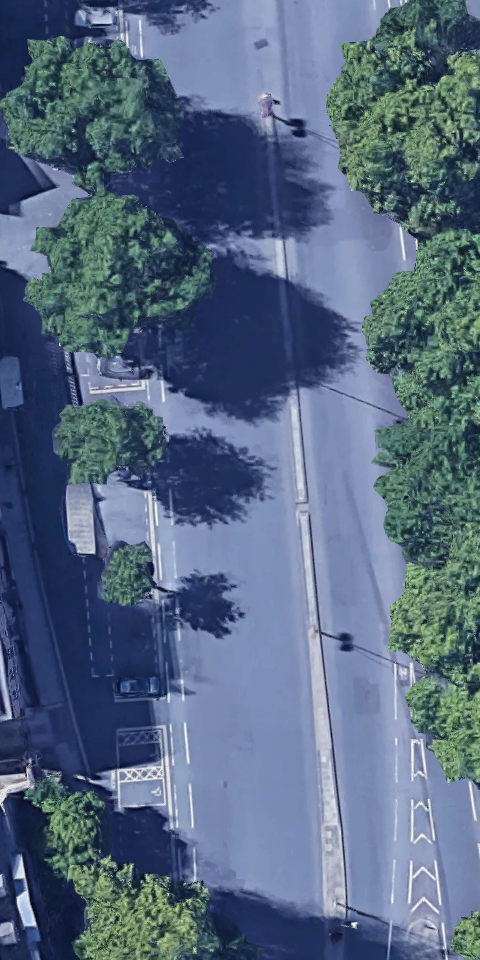}
    
    \vspace{0.1em}
    
    \includegraphics[width=0.195\textwidth]{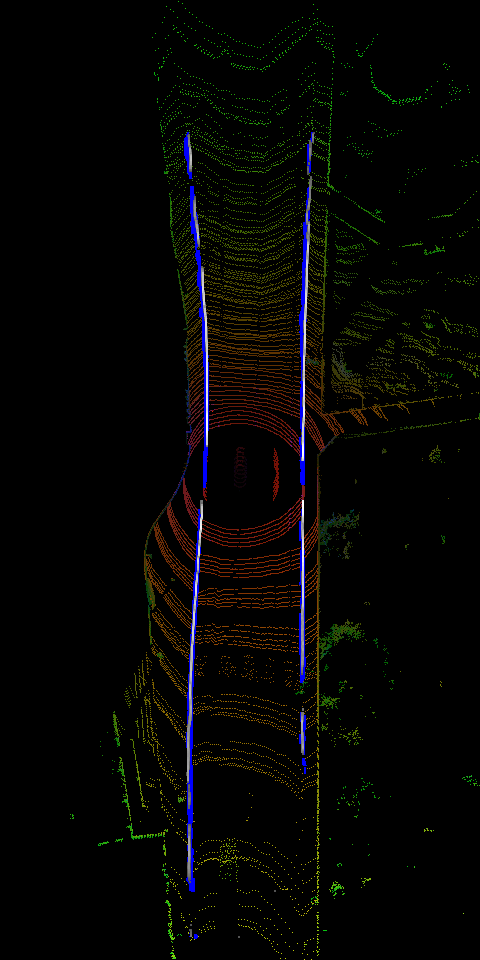}
    \includegraphics[width=0.195\textwidth]{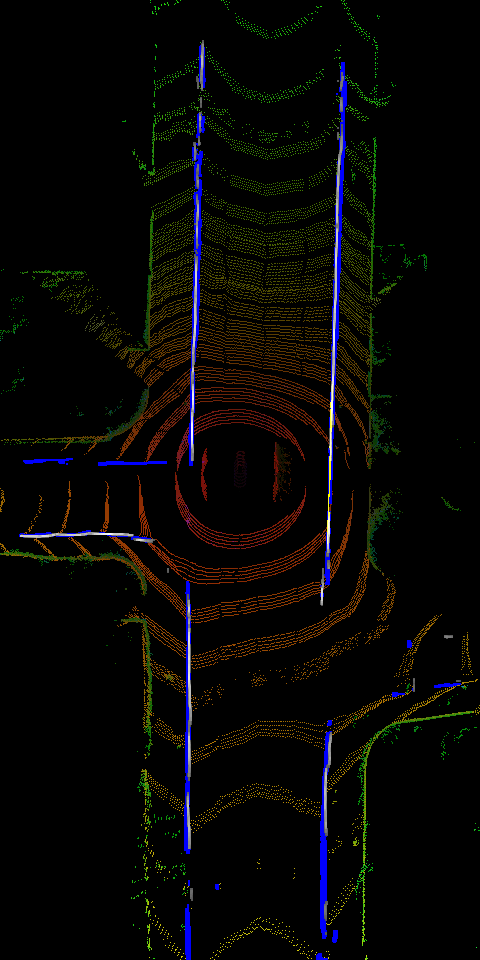}
    \includegraphics[width=0.195\textwidth]{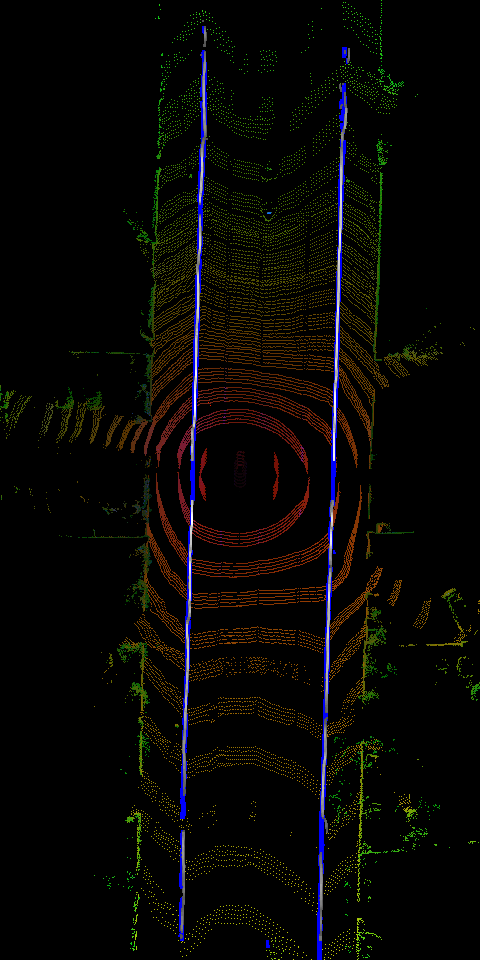}
    \includegraphics[width=0.195\textwidth]{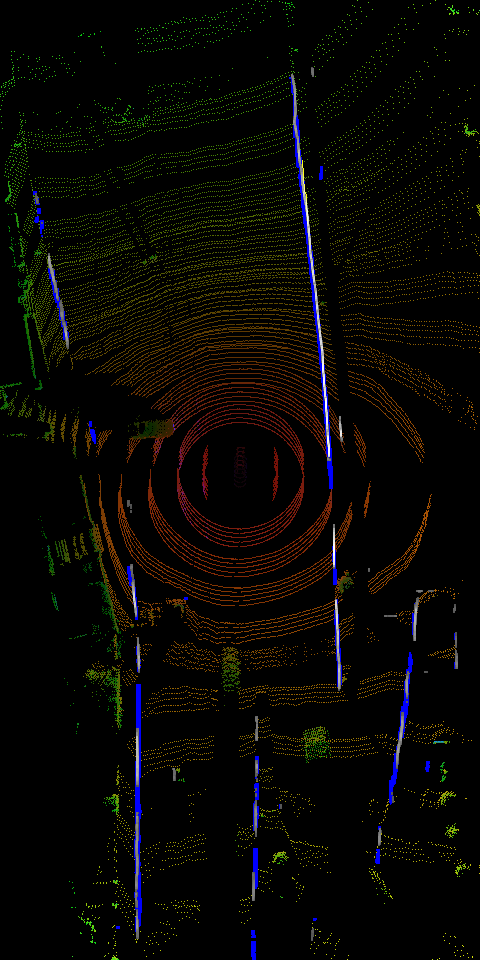}
    \includegraphics[width=0.195\textwidth]{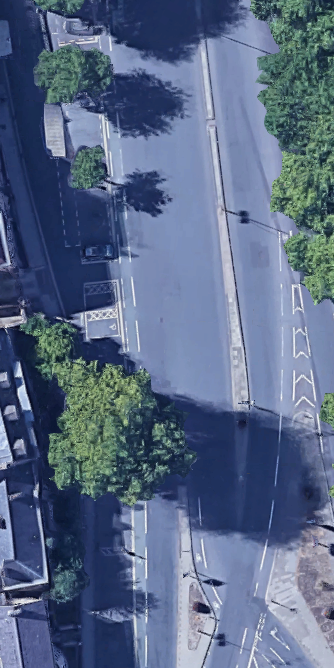}
    \caption{First row: Sample outputs from the networks of detected and inferred road boundaries.
    Second row: Sample outputs after post-processing steps. The satellite images on the right depict the complexity of the scenes next to them.}
    \label{fig:output_examples}
\end{figure*}


\section{Experimental Results}
\label{sec:experimental_results}

In this section, we show some qualitative results and provide an extensive quantitative evaluation of our approach. Qualitative results are presented in Figure \ref{fig:output_examples} where outputs of curb detection networks and post-processing steps are included. The examples show that our approach is able to generate accurate masks for detected visible and inferred occluded road boundaries. To the best of our knowledge there is no public road boundary detection benchmark. Hence, we could not undertake a quantitative comparison of our approach with other existing approaches. However, we present an extensive quantitative evaluation based on our ground truth data.

\subsubsection{Visible only} as we described in Section \ref{sec:approach}, our approach tackles the detection of visible and inference of occluded road boundaries in two steps with a continuity constraint. The visible detector model is trained with 21K samples, which are generated by augmenting 1300 bird's-eye view images. Table \ref{tab:visible_only_accuracy} summarises the accuracy of the model with respect to 500 test images. We present precision, recall and F1 score for different areas around the vehicle (images size) and different tolerances. With minimum tolerance and maximum distance of 96 metres the model achieves 0.8168 F1 score. The model performs better in areas close to the car and achieves up to 0.9437 F1 score.

\begin{table}[t!]
\centering
\caption[]{Precision, recall and F1 score of the visible road boundary detection model}
\resizebox{\columnwidth}{!}{%
\begin{tabular}{@{}lclll@{}}
\toprule
Image size / area & Tolerance & Precision & Recall & F1 Score \\
\midrule
\multirow{4}{*}{480x960 / 48x96 metres} & 1px & 0.8371 & 0.7974 & 0.8168 \\
    & 2px & 0.9193 & 0.8387 & 0.8772 \\
    & 3px & 0.9455 & 0.8572 & 0.8992 \\
    & 4px & 0.9579 & 0.8698 & 0.9117 \\
\midrule
\multirow{4}{*}{480x720 / 48x72 metres} & 1px & 0.8631 & 0.8721 & 0.8676 \\
    & 2px & 0.9310 & 0.9050 & 0.9178 \\
    & 3px & 0.9520 & 0.9183 & 0.9349 \\
    & 4px & 0.9618 & 0.9272 & 0.9442 \\
\midrule
\multirow{4}{*}{480x480 / 48x48 metres} & 1px & 0.8819 & 0.8921 & 0.8870 \\
    & 2px & 0.9343 & 0.9021 & 0.9179 \\
    & 3px & 0.9541 & 0.9157 & 0.9345 \\
    & 4px & 0.9632 & 0.9250 & 0.9437 \\
\bottomrule
\end{tabular}
\label{tab:visible_only_accuracy}}
\end{table}

\subsubsection{Visible and occluded} the occluded road boundary inference model is also trained with 21K samples that are obtained from 600 bird's-eye images through augmentation. The model infers road boundaries that are occluded by other road users and achieves 0.7867 F1 score. Note that for occluded road boundaries having higher tolerance is acceptable as their exact location is unknown. Combining outputs of both networks gives 0.893 F1 score with respect to all road boundaries.

\begin{table}[t!]
\centering
\caption[]{Precision, recall and F1 score of the visible and occluded road boundary detection and inference models (size: 480x960)}
\resizebox{\columnwidth}{!}{%
\begin{tabular}{@{}llclll@{}}
\toprule
Model & Labels & Tolerance & Precision & Recall & F1 Score \\
\midrule
\multirow{4}{*}{Occluded} & \multirow{4}{*}{Occluded} & 1px & 0.5099 & 0.6420 & 0.5684 \\
    &  & 2px & 0.6414 & 0.7441 & 0.6889 \\
    &  & 3px & 0.7090 & 0.7938 & 0.7490 \\
    &  & 4px & 0.7523 & 0.8245 & 0.7867 \\
\midrule
\multirow{4}{*}{Vis + Occ} & \multirow{4}{*}{Vis + Occ} & 1px & 0.8318 & 0.7531 & 0.7905 \\
    &  & 2px & 0.9184 & 0.7980 & 0.8540 \\
    &  & 3px & 0.9477 & 0.8191 & 0.8787 \\
    &  & 4px & 0.9619 & 0.8333 & 0.8930 \\
\bottomrule
\end{tabular}
\label{tab:visible_occluded_accuracy}}
\end{table}

\subsubsection{Post-processing}

The importance of the post-processing step is to track detected and inferred road boundaries and fill in the gaps that were not detected initially. This happens on both sides of the vehicle as can be seen in Figure \ref{fig:post_processing_steps} (left). Those gaps are closed after the post-processing step (right). Note that when generating filtered outputs ($\Upsilon_i$) dilation is applied to the output masks to increase the overlap between output masks. As a result, the thickness of detected and inferred road boundaries increases, which leads to a decrease in precision and F1 score but an increase in recall for the minimum tolerance. However, whenever the tolerance is above 1px, the F1 score increases as the post-processing step fills in the gaps. 

\begin{table}[t!]
\centering
\caption[]{Precision, recall and F1 score with and without post-processing step (size: 480x480)}
\resizebox{\columnwidth}{!}{%
\begin{tabular}{@{}lclll@{}}
\toprule
Pipeline & Tolerance & Precision & Recall & F1 Score \\
\midrule
\multirow{4}{*}{Vis + Occ} & 1px & 0.8922 & 0.8387 & 0.8646 \\
    & 2px & 0.9470 & 0.8649 & 0.9041 \\
    & 3px & 0.9627 & 0.8784 & 0.9186 \\
    & 4px & 0.9706 & 0.8887 & 0.9279 \\
\midrule
\multirow{4}{*}{Vis + Occ + Post} & 1px & 0.5961 & 0.9246 & 0.7249 \\
    & 2px & 0.8773 & 0.9524 & 0.9133 \\
    & 3px & 0.9542 & 0.9594 & 0.9568 \\
    & 4px & 0.9741 & 0.9630 & 0.9685 \\
\bottomrule
\end{tabular}
\label{tab:visible_only_accuracy}}
\end{table}

\subsubsection{Running time}

The system runs at 8.53 Frames Per Second (FPS) with input images of size 480x960 on a NVIDIA 1080 Ti GPU.

Overall, the experiments demonstrate that our approach is capable of detecting and inferring all road boundaries in complex road scenes over a total distance of 96 metres around the vehicle and in real-time.


\section{Conclusions}
\label{sec:conclusions}

In this paper, we have presented a LIDAR-based approach for curb detection around the vehicle. Integrated LIDAR pointclouds are transformed into bird's-eye images which are then processed by trained convolutional networks. Instead of processing the whole pointcloud we only process a projected pointcloud in form of an image which allows our method to operate in real-time. Finally, we post-process the network outputs, i.e. the detected curbs (visible and occluded), by filtering out noise and  tracking detections over time. 
This last step increases the overall performance of curb detection as we have shown in an extensive evaluation (Section~\ref{sec:experimental_results}). 

Overall, we conclude that LIDAR data can be used to accurately detect curbs at high processing rates. Compared to camera-based based approaches, LIDAR has not only a greater coverage and range around the vehicle, but it is also more robust to environmental changes due to lighting and/or weather. The experiments demonstrated that our approach achieves a high performance in detecting and inferring visible and occluded road boundaries around the vehicle and achieves an F1 score of 0.9685 with post-processing. Hence, we strongly believe that our proposed deep learning approach based on LIDAR data can have a wide impact on a range of different applications in the autonomous driving domain.

\paragraph*{Acknowledgment}
The work has been supported by the EPSRC/UK Research and Innovation Programme Grant EP/M019918/1 (Mobile Autonomy) and the NVIDIA Corporation with the donation of Titan Xp and Titan V GPUs.



\end{document}